\RequirePackage[T1]{fontenc}
\documentclass[conference]{IEEEtran}
\usepackage{amsmath,amssymb,booktabs,array,graphicx}
\usepackage{tikz}
\usetikzlibrary{arrows.meta,positioning,shapes.geometric}
\usepackage[hidelinks]{hyperref}
\usepackage{url}
\usepackage{cite}
\newcommand{\code}[1]{\texttt{#1}}

\begin{document}
\title{Bengali Hallucination Detection with Retrieval and Programs: A System Study}
\author{
\IEEEauthorblockN{1\textsuperscript{st} Shamim Hasan}
\IEEEauthorblockA{\textit{Department of Computer Science and Engineering} \\
\textit{BNIST}\\
Dhaka, Bangladesh \\
adilshamim696@gmail.com}
\and
\IEEEauthorblockN{2\textsuperscript{nd} Samiya Ahmed Chowdhury} 
\IEEEauthorblockA{\textit{Department of Computer Science and Engineering} \\
\textit{Chittagong University of Engineering and Technology}\\
Chittagong, Bangladesh \\
samiya.ahmed.chowdhury@gmail.com}
}

\maketitle
\begin{abstract}
Bengali hallucination detection must distinguish unsupported factual claims from incorrect reasoning, often without a supplied reference passage. We present and characterize an inference-only system that combines response-conditioned query rewriting, hybrid retrieval, and task-specific language-model judging. Six retrieval routes search question--answer and passage collections; a router directs each example to a 27B textual judge or a 14B agent that can compare retrieved answers, reason directly, or generate and execute Python. We characterize the system using version-specific code, saved predictions, execution traces, and corpus metadata from the Olikbochon challenge. Kaggle associates a displayed private macro-F1 of 0.956 with Version 2; its saved run processes 2,516 examples in 220.99 minutes on two T4 GPUs. Of 253 math-routed examples, 88 terminate through candidate matching, while execution traces expose 49 stderr-bearing run events across 35 examples. A later run changes six predictions despite identical core worker code. A bounded audit of 1,090 benchmark-related corpus records finds exact question or context overlap for 38 processed evaluation inputs, precluding an unqualified claim of corpus-independent generalization. The contribution is a concrete retrieval-and-computation judging architecture with an auditable operational characterization, not a controlled estimate of component gains. We distinguish the reported leaderboard result from independently recomputed diagnostics and identify the evidence needed for stronger effectiveness and reproducibility claims.
\end{abstract}

\section{Introduction}
Fluent language generation does not establish factual correctness. In Bengali, a detector may need to verify a locally specific fact, interpret an idiom, compare a summary with a supplied passage, or assess a mathematical derivation. These settings differ in where valid evidence can be obtained. A passage can establish source faithfulness, an external collection can supply factual references, and a computation can test a numerical claim. Treating all three as the same question to a language model obscures these differences.

The Olikbochon Bengali LLM Hallucination Detection Challenge formalizes this problem as binary classification of a candidate response given a Bengali prompt and optional context~\cite{iutcs2026bengalihallucination}. It releases a small labeled sample rather than a large conventional training split. This motivates investigating inference-time composition of pretrained models and external resources. It does not, however, make such a system zero-resource: the reference corpus, prompts, and model checkpoints remain substantive dependencies.

We document our publicly released competition implementation~\cite{shamim2026notebook}. Its main architectural choice is to separate evidence acquisition from judgment and to distinguish lookup-oriented questions from derivation-oriented questions. A small model rewrites the candidate claim for retrieval; multilingual dense and lexical searches retrieve candidate references; a 27B judge handles non-math items; and a 14B agent uses retrieved-answer matching or program-assisted reasoning on math-routed items. Models are used without task-specific weight updates.

The design assigns distinct roles to evidence sources: question-only routes seek topic matches, response-conditioned routes seek the candidate claim, and context routes recover passage-level evidence. The router separates lookup from derivation, while sequential model loading accommodates the two-GPU setting. We pair this design with a version-resolved analysis of saved artifacts; the individual benefits of these choices are not measured.

Our contributions are threefold:
\begin{enumerate}
\item A precise account of a Bengali judging pipeline integrating response-conditioned hybrid retrieval, a textual judge, and a lookup/reasoning/program-execution agent under a two-GPU inference configuration.
\item A reproducible operational characterization of two saved runs, including disjoint route counts, model-stage timings, execution failures, terminal defaults, and cross-version prediction changes.
\item A bounded provenance and overlap analysis that qualifies the interpretation of the reported leaderboard result and separates verified observations from untested causal explanations.
\end{enumerate}
The retrieval, program-aided reasoning, and serving techniques are established. The contribution is a system characterization, not algorithmic novelty or measured component-level gains.

\section{Related Work}
\subsection{Hallucination Evaluation and Bengali Resources}
HaluEval evaluates hallucination recognition across question answering, dialogue, and summarization~\cite{li-etal-2023-halueval}. FActScore decomposes generated text into atomic claims and measures factual precision against a knowledge source~\cite{min-etal-2023-factscore}. The present task differs from long-form factual precision: it assigns one label to an entire candidate response, and may provide a governing context. Multilingual evaluation further requires care with task construction and estimation protocols~\cite{ul-islam-etal-2025-much}; an English benchmark score does not establish Bengali detection quality.

BanglaBERT demonstrates the value of language-specific pretraining and evaluation resources for Bengali understanding~\cite{bhattacharjee-etal-2022-banglabert}. BenHalluEval studies Bengali hallucination evaluation across multiple tasks~\cite{adib2026benhalueval}. The competition host states that portions of Phase 1 follow its methodology, while a later held-out phase is intended to exclude samples from the associated datasets~\cite{phase2preview}. This establishes context, not identity of the benchmark datasets. We do not compare BenHalluScore numerically with competition macro-F1: they are different measures on different evaluation protocols.

\subsection{Retrieval and Program-Assisted Reasoning}
Retrieval-augmented generation conditions predictions on external text~\cite{lewis2020rag}. HyDE generates hypothetical documents to improve dense retrieval without relevance labels~\cite{gao-etal-2023-precise}. Our rewrite is related but materially different: it is conditioned on the response being evaluated and is instructed to preserve that response's claim, even if false. It is therefore described as \emph{response-conditioned rewriting}, rather than generation of a hypothetical correct answer.

The retrieval stack combines multilingual E5 embeddings~\cite{wang2024multilinguale5}, FAISS search~\cite{johnson2021faiss}, BM25~\cite{robertson2009bm25}, reciprocal rank fusion (RRF)~\cite{cormack2009rrf}, and the BGE-reranker-v2-m3 cross-encoder~\cite{baaiBgeRerankerV2M3}. These components provide complementary matching mechanisms, but their combination is not itself evidence of improved downstream detection. In particular, reranker relevance scores should not be interpreted as calibrated probabilities of factual correctness.

Program-aided language models~\cite{gao2023pal} and Program of Thoughts~\cite{chen2023programthoughts} separate parts of numerical reasoning from computation. Our MathAgent applies this principle to judging a candidate answer, with a retrieved-answer shortcut and a model-based verification step. Executable code can expose arithmetic and symbolic errors, but it cannot guarantee that the generated program formalizes the question correctly. Qwen models~\cite{qwen3technicalreport,qwen35_27b_modelcard}, AWQ quantization~\cite{lin2024awq}, and vLLM serving~\cite{kwon2023pagedattention} provide the pretrained and inference infrastructure, rather than independently novel contributions of this work.

Tool-augmented factuality checking is directly relevant. FacTool combines claim/query generation, external evidence collection, and tool-based verification across knowledge-based QA and mathematical reasoning~\cite{chern2023factool}. We do not introduce that general combination. Our system instantiates response-level Bengali judging with a local hybrid-retrieval package, a retrieved-answer shortcut, and staged open-weight inference. These are design differences, not demonstrated advantages over FacTool.

\section{Task, Data, and Evaluation Protocol}
\subsection{Problem Formulation}
An input is $x_i=(q_i,r_i,c_i)$, where $q_i$ is the prompt, $r_i$ the candidate response, and $c_i$ an optional context. The target is $y_i\in\{0,1\}$, with 0 denoting hallucinated and 1 faithful. The system produces
\begin{equation}
\hat y_i=f_{\theta,\mathcal D}(q_i,r_i,c_i),
\end{equation}
where $\theta$ denotes fixed pretrained models and inference policies, and $\mathcal D$ is the external retrieval collection. No training loss is optimized in the notebook. A response may be judged against the supplied context or external evidence, depending on task and source relevance; faithfulness to a source is not identical to world truth.

\subsection{Released Data and Competition Scope}
The official host starter reports 299 labeled examples: 136 hallucinated and 163 faithful, with 130 containing context~\cite{sushmit2026starter}. The saved pipeline run loads 2,516 test examples with fields \code{id}, \code{context}, \code{prompt\_bn}, and \code{response\_bn}. Its context normalization identifies 1,361 context-bearing and 1,155 context-absent examples. Test labels are not available in the artifacts analyzed here. Table~\ref{tab:data} separates labeled-sample statistics from unlabeled test coverage; prediction balance must not be substituted for test class balance.

\begin{table}[t]
\caption{Available data. Context counts follow the respective published normalization procedures. Test class counts are unavailable.}
\label{tab:data}\centering
\begin{tabular}{lrr}
\toprule
Property & Labeled sample & Test inputs\\
\midrule
Examples & 299 & 2,516\\
Hallucinated / faithful & 136 / 163 & Not released\\
With context & 130 & 1,361\\
Without context & 169 & 1,155\\
\bottomrule
\end{tabular}
\end{table}

The overview distinguishes globally stable facts (C0), culturally situated knowledge (C1), and contested or time-sensitive facts (C2). It also describes task types ranging from closed-book QA to translation. Band annotations and the private split membership are not available here, so neither band-wise F1 nor out-of-distribution performance can be estimated. The code reads labeled samples for exploratory analysis; it does not evaluate the complete pipeline on them or train a detector from them.

Competition rules allow public external data and fine-tuning. The later reproduction phase requires offline inference with open-weight models, a P100 or two T4 GPUs, less than nine hours, and under 50 GB of model weights~\cite{competitionrules}. The audited notebook runs with Internet enabled for package/model acquisition. Its hardware and observed duration fall within two stated limits, but this alone does not demonstrate offline reproduction or satisfaction of every rule. A Phase 1 private leaderboard score is also distinct from the event's combined final score and its new held-out evaluation.

\subsection{Metric and Evidence Boundaries}
We use the host's explicit clarification of unweighted class-macro F1~\cite{metricclarification}, resolving inconsistent wording elsewhere on the competition pages:
\begin{equation}
F_{1,k}=\frac{2\mathrm{TP}_k}{2\mathrm{TP}_k+\mathrm{FP}_k+\mathrm{FN}_k},\qquad
F_{1,\mathrm{macro}}=\frac{F_{1,0}+F_{1,1}}{2}.
\label{eq:f1}
\end{equation}
A zero denominator requires the evaluator's convention; it is not inferable from saved predictions. The displayed score is rounded to three decimals. Without gold labels and the private split, we cannot recompute it, derive a confusion matrix, or attach a valid confidence interval. In particular, $1-F_{1,\mathrm{macro}}$ is not an error rate.

\section{Methodology}
\subsection{Pipeline and Text Processing}
\begin{figure*}[t]
\centering
\begin{tikzpicture}[>=Stealth, every node/.style={font=\small}, box/.style={draw,rounded corners=1pt,align=center,text width=3.3cm,minimum height=1.05cm,inner sep=5pt}, flow/.style={->,thick}]
\node[box] (input) at (0,0) {Prompt, response,\\optional context};
\node[box] (norm) at (4,0) {Text processing\\and 4B claim rewrite};
\node[box] (retr) at (8,0) {Six hybrid routes\\RRF + rerank $\to$ top 5};
\node[box] (router) at (12,0) {14B question-only\\math router};
\draw[flow] (input)--(norm);
\draw[flow] (norm)--(retr);
\draw[flow] (retr)--(router);
\node[box,text width=5.3cm,minimum height=1.25cm] (text) at (4,-2.1) {27B textual judge (TP=2)\\High-score or decreasing-evidence path\\Supplied context + retrieved records};
\node[box,text width=5.3cm,minimum height=1.25cm] (math) at (11,-2.1) {14B MathAgent\\Candidate match / reasoning / Python\\Question + response + retrieved records};
\draw[flow] (router.south)--++(0,-0.35)-|node[pos=0.32,above]{non-math} (text.north);
\draw[flow] (router.east)--++(0.5,0)|-node[pos=0.65,right]{math} (math.east);
\node[box,text width=5.3cm,minimum height=0.9cm] (out) at (7.5,-4) {Parsed yes/no $\to$ binary submission\\Unresolved decisions default to faithful};
\draw[flow] (text.south)|-(out.west);
\draw[flow] (math.south)|-(out.east);
\node[align=center,font=\footnotesize] (retry) at (0.5,-3.1) {Short verdict\\format retry};
\draw[flow,dashed] (text.west)-|(retry.north);
\draw[flow,dashed] (retry.south)|-(out.west);
\end{tikzpicture}
\caption{Executed architecture. The textual judge receives supplied context, but the MathAgent payload does not. High-score gates use rounded reranker values, not verified exact matches. Dashed branches indicate fallback behavior rather than independent verification.}
\label{fig:architecture}
\end{figure*}
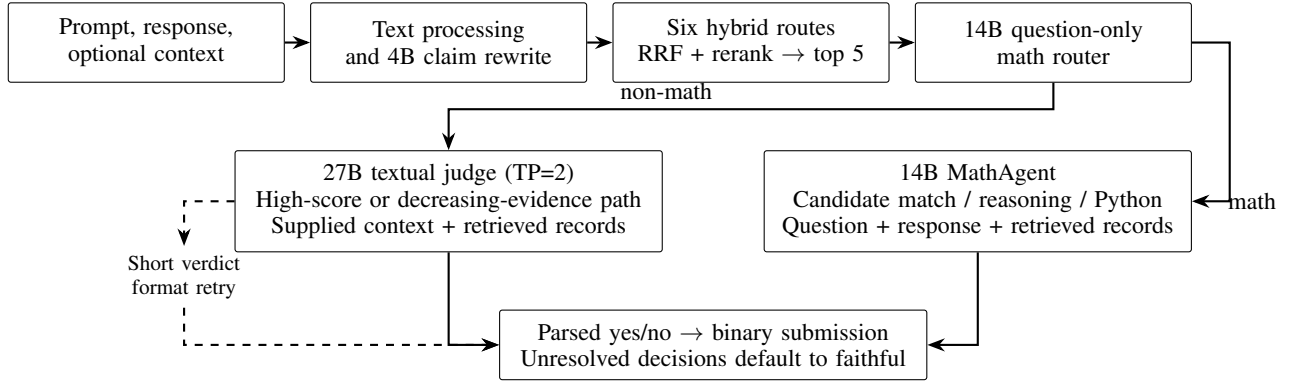

Figure~\ref{fig:architecture} shows the executed structure. Retrieval precedes routing and therefore processes both math and non-math inputs. Text processing removes HTML/SVG-like fragments, URLs and email-like strings, selected script ranges, and characters outside an allowlist; it then collapses whitespace. Bengali characters and numerals, ASCII text, and selected mathematical symbols remain. This is not lossless Unicode normalization: the same transformation is applied to the response being judged and can remove meaningful content. No general numeral-to-Python transliteration function is implemented; code generation must express usable numerical literals itself.

\subsection{Response-Conditioned Hybrid Retrieval}
Let $N$ denote text normalization and $h_i=G_4(N(q_i),N(r_i))$ the 4B model's declarative rewrite. Its prompt asks it to insert the candidate response into a statement without correcting or inferring facts. The rewrite is a retrieval query, not evidence. A false response can consequently influence which evidence the judge receives.

Six routes query either question-only Q\&A representations or context passages (Table~\ref{tab:routes}). Empty queries are skipped, and the last two routes require a nonempty supplied context. For each route, E5 encodes a \code{query:} prefixed query with normalized embeddings. FAISS and BM25 each return up to 20 candidates. RRF combines the two ranked lists:
\begin{equation}
\mathrm{RRF}(d)=\sum_{L\in\{L_{\rm dense},L_{\rm sparse}\}:d\in L}
\frac{1}{60+\operatorname{rank}_{L}(d)},
\label{eq:rrf}
\end{equation}
with ranks starting at 1. The 20 highest fused candidates are reranked as query--text pairs, and five are retained per route. For Q\&A records, the reranker sees the question text, not the stored answer.

\begin{table}[t]
\caption{Retrieval routes. $\tilde q=N(q)$ and $\tilde c=N(c)$; $h$ is the response-conditioned rewrite.}
\label{tab:routes}\centering
\begin{tabular}{cll}
\toprule
Route & Query & Collection\\
\midrule
A & $\tilde q$ & Q\&A questions\\
B & $h$ & Q\&A questions\\
C & $\tilde q$ & Context passages\\
D & $h$ & Context passages\\
E & $\tilde c$, if present & Context passages\\
F & $\tilde c\,\Vert\,\tilde q$, if present & Context passages\\
\bottomrule
\end{tabular}
\end{table}

The union contains at most 30 candidates before deduplication. Records are deduplicated by collection kind and ID, retaining the highest reranker score and breaking ties by route order. The global top five form $K_i$. Scores from different query formulations are compared directly, without cross-query calibration. Let $s_i(d)$ be the maximum unrounded score among retained route copies of record $d$. Downstream scores are serialized to four decimal places; the high-score gate is
\begin{equation}
H_i=\{d\in K_i:\operatorname{round}(s_i(d),4)\geq 1\}.
\label{eq:gate}
\end{equation}
Neither this threshold nor the code's historical ``exact'' label establishes textual identity, semantic equivalence, or truth.

The executed logs resolve the final IUFree dataset~\cite{nafiz2026corpus}. Its version-1 Q\&A and context stores contain 391,762 and 1,038,167 records, respectively. An additional 391,762-record QA representation is packaged but not searched by these six routes; it must not be counted as independent evidence. Source labels span encyclopedic, dictionary, idiom, textbook, news, law, and QA material. These counts describe stored records, not unique documents. The package supplies a shard-embedding worker and index metadata, but not a complete upstream acquisition and decontamination recipe.

\subsection{Routing and Textual Judging}
The 14B router receives only the normalized question. It distinguishes answers that must be derived from facts that can be looked up; a numeric answer does not alone imply mathematics. The active configuration uses one greedy generation, not majority voting. A malformed routing output defaults to non-math.

For non-math inputs with $H_i\neq\varnothing$, the 27B judge receives up to three high-score records and any supplied context. Its policy gives the supplied context priority when it addresses the question. Otherwise a matching retrieved question or relevant supporting record governs the decision. It accepts meaning-equivalent paraphrases, while additions or contradictions can produce a hallucination verdict. The policy tells the judge not to second-guess the governing source; a wrong source can therefore support a wrong verdict.

Other non-math inputs follow a decreasing-evidence schedule of three, two, then one retrieved records, with supplied context included where available. Retries continue only while no parseable verdict is produced; they are not confidence-based escalation or a consistency ensemble. This path permits parametric knowledge when provided evidence is irrelevant. High-score-path failures do not enter the decreasing-evidence path. Both paths can receive a final short verdict-format retry. Inputs with neither context nor retrieved records are omitted from the latter path and reach the terminal default. Thus the implementation mixes source-based judgment and model knowledge rather than enforcing pure entailment for every example.

\subsection{Lookup and Program-Assisted MathAgent}
For math-routed inputs, a 14B agent first inspects up to three high-score candidates, drawn from either collection. It checks whether a candidate has the same question objective, arguments, and numbers. If it finds a match and emits a verdict, it trusts the stored answer and terminates without solving the problem. This shortcut is answer comparison, not independently verified mathematics.

Otherwise a triage step selects non-math reconsideration, direct reasoning, or code-assisted solving. The code path produces a strategy and Python program, executes the program, and asks the same model to inspect the code/output and compare the candidate response with its derived answer. Execution failures may trigger repair, with at most five code executions and a requested 20-second timeout per execution. If code cannot be produced or the process cannot resolve the item, a model-only fallback may be used. SymPy is requested by prompts but is not enforced. Neither a successful process exit nor the verifier's \code{exact\_answer} field proves symbolic correctness.

The MathAgent payload omits supplied context. In Version 2, one of 253 math-routed rows has usable context (ID 2338); its label changes from 0 in Version 2 to 1 in Version 3. This identifies an exposed case, not an error or a causal effect of context omission. All 157 Version 2 program-path rows attempt execution; 155 contain a verifier-judge event and two terminate through textual fallback after attempts. Neither event establishes mathematical correctness.

\subsection{Aggregation and Compute Orchestration}
The parsed internal decision is $v_i\in\{\mathrm{yes},\mathrm{no},\mathrm{unknown}\}$, with submission mapping
\begin{equation}
\hat y_i=\begin{cases}0,&v_i=\mathrm{no},\\1,&\text{otherwise}.\end{cases}
\label{eq:default}
\end{equation}
Malformed or missing decisions also map to 1. This is a faithful-class default, not an abstention mechanism or a conservative safety policy.

The 4B and 14B stages use independent workers on each GPU; the 27B stage uses one tensor-parallel model across both GPUs. Model stages run in separate processes and exchange local files. Process exit releases model allocations before the next stage loads. This is sequential model orchestration, not simultaneous residency of all models, and not a novel parallelism method. Generated Python also runs in separate processes, but with unrestricted builtins/imports. The timeout and process boundary do \emph{not} constitute a security sandbox.

\section{System Configuration and Analysis Protocol}
\subsection{Version-Resolved Evidence}
We analyze the exact executed notebooks and saved outputs for Version 2 (script ID 340370414) and Version 3 (347400763), retrieved on September 5, 2026~\cite{shamim2026notebook}. Kaggle's public best-score link explicitly identifies Version 2 for 0.956. We therefore associate that displayed score only with Version 2, and use its predictions for the primary descriptive analysis. The saved Version 2 execution is dated August 5, 2026, after the July 19 close displayed in the overview. We did not verify deadline extensions or recover a receipt linking the downloaded submission hash to the scored entry. The badge establishes a platform-level version association, not contemporaneous submission identity.

The five generated worker scripts are byte-identical across these versions; source changes are largely explanatory and diagnostic. The execution environments nevertheless differ: saved logs identify vLLM 0.26.0 in Version 2 and 0.28.0 in Version 3. We do not rerun GPU inference. Instead, we recompute counts from CSV outputs and inspect JSON event traces without deserializing executable pickle files. The supplementary package reproduces saved-output counts, trace diagnostics, worker-byte comparisons, selected cell timings, and cross-version label differences, with versioned retrieval instructions and input hashes.

\subsection{Active Configuration}
The active generative checkpoints are \path{Qwen/Qwen3-4B-AWQ}, \path{Qwen/Qwen3-14B-AWQ}, and \path{QuantTrio/Qwen3.5-27B-AWQ}; the latter is a third-party quantized checkpoint~\cite{quanttrio2026awq}, not the base-model card itself. Retrieval uses \path{intfloat/multilingual-e5-large} and \path{BAAI/bge-reranker-v2-m3}. Table~\ref{tab:config} reports consumed settings rather than every declared variable. In particular, the notebook declares retrieval values of 50 that do not control the active worker; the actual pool is 20.

\begin{table}[t]
\caption{Selected active settings, verified against executable worker code. Generation budgets count new tokens.}
\label{tab:config}\centering\small
\begin{tabular}{p{0.62\columnwidth}r}
\toprule
Setting & Value\\
\midrule
Dense / sparse candidates per route & 20 / 20\\
RRF constant / rerank pool / global retention & 60 / 20 / 5\\
Embedding / reranker token limits & 512 / 512\\
Embedding / reranker batch sizes & 256 / 768\\
Rewrite / router generation budgets & 100 / 8\\
27B judgment / format-retry budgets & 120 / 6\\
27B maximum model length & 5,000\\
27B tensor parallelism / GPU utilization & 2 / 0.96\\
MathAgent maximum model length & 4,096\\
Math candidate-match budget & 256\\
Maximum code executions / timeout & 5 / 20 s\\
\bottomrule
\end{tabular}
\end{table}

All active generation uses temperature zero. Greedy decoding does not ensure identical outputs across library versions, hardware kernels, or batching. Package installations and model revisions are not fully pinned; the external index build is also incompletely specified. The saved successful two-T4 runs demonstrate feasibility of those recorded executions, not a guarantee that current dependency resolution reproduces them.

\subsection{Analyses and Their Estimands}
We measure route coverage, verdict/default frequencies, event-level execution problems, elapsed cell times, and cross-version disagreement. Each is an operational statistic, not an accuracy estimate. Timing comes from Papermill's per-cell and notebook metadata and includes initialization, model loading, and applicable setup; it is not isolated token-generation latency.

For provenance analysis, we inspect the final corpus's metadata and selected source columns. A bounded exact-overlap audit compares 1,090 selected records---77 Q\&A and 88 context records labeled \code{comp\_original}, one \code{banglahall\_qna} record, and 924 \code{banglahall\_context} records---against saved Version 2 inputs. These categories were selected by source name rather than randomly sampled. Comparisons use already normalized question/context strings saved by the pipeline, not independently retrieved original competition text. Every match also holds before the audit's additional whitespace collapse. Thus exactness refers to processed-input equality. We separately inspect retained retrieval slots; original-input identity, near duplicates, hidden-label overlap, and full-corpus decontamination are not established.

\section{Results}
\subsection{Reported Score and Saved Route Coverage}
Kaggle displays private macro-F1 \textbf{0.956} for Version 2. The public evidence does not expose the unrounded value, private labels, per-class F1, or score-bearing receipt. Accordingly, the result is reported with its provenance rather than reconstructed from a hard-coded notebook annotation. The annotation itself is not evaluation code. No comparable score for the saved Version 3 predictions was independently verified. The following diagnostics cover all 2,516 inputs, not the undisclosed private scoring subset.

Table~\ref{tab:stages} gives disjoint terminal-stage counts. The router sends 253/2,516 examples (10.06\%) to the MathAgent. The high-score textual path covers 1,991 examples (79.13\%), including one format retry; another 272 finish on the first three-record general path. No examples terminate in the two- or one-record retry stages in these saved runs. The absence of those terminal stages is not evidence of a bug, because earlier outputs may already be parseable.

\begin{table}[t]
\caption{Disjoint terminal-stage counts from saved outputs. ``High-score'' replaces the code's misleading ``exact'' terminology. Program-path counts include model-only fallbacks.}
\label{tab:stages}\centering\small
\begin{tabular}{p{0.67\columnwidth}rr}
\toprule
Terminal route & V2 & V3\\
\midrule
High-score textual, no supplied context & 755 & 755\\
High-score textual, with context & 1,235 & 1,235\\
High-score textual, format fallback & 1 & 1\\
General top-three, no context & 148 & 148\\
General top-three, with context & 124 & 124\\
Math candidate match & 88 & 89\\
Math program-path category & 157 & 158\\
Math direct reasoning & 6 & 6\\
Math non-math reconsideration & 2 & 0\\
\midrule
Total & 2,516 & 2,516\\
\bottomrule
\end{tabular}
\end{table}

Within the 253 math-routed examples in Version 2, 88 (34.78\%) terminate through candidate matching. This shows that the shortcut is used, not that it improves F1 or that its matches are all correct. The final internal decisions are 1,291 yes, 1,224 no, and one unknown. After Eq.~\eqref{eq:default}, the submission contains 1,292 faithful and 1,224 hallucinated labels. These are model decisions, not a recovered test-label distribution.

\subsection{Execution Cost and Cross-Version Variation}
Version 2's Papermill duration is 13,259.21 s (220.99 min). Its joint 27B judging cell consumes 9,433.41 s (157.22 min), approximately 71.15\% of total notebook time. Table~\ref{tab:time} shows the recorded breakdown. Both the substantial retrieval cost and dominant 27B cost matter for a deployment decision; an aggregate batch runtime is not a per-request latency measurement.

\begin{table}[t]
\caption{Recorded elapsed minutes, including initialization within each cell. Total includes all notebook cells and overhead; listed stages need not sum to total. Runs differ in environment and are not a controlled speed comparison.}
\label{tab:time}\centering
\begin{tabular}{lrr}
\toprule
Operation & V2 & V3\\
\midrule
Dependency installation & 4.52 & 4.51\\
Claim rewriting & 7.64 & 7.02\\
Retrieval and reranking & 31.84 & 28.98\\
Math routing & 7.21 & 6.71\\
Joint 27B judging & 157.22 & 147.44\\
MathAgent & 11.96 & 11.08\\
\midrule
Full notebook & 220.99 & 206.33\\
\bottomrule
\end{tabular}
\end{table}

Version 3 takes 12,379.76 s (206.33 min), but has a different vLLM version and is not an isolated optimization experiment. The saved routing CSV is byte-identical between runs; six final labels differ (0.2385\%). Four change from faithful to hallucinated and two change in the opposite direction. The result demonstrates output variation under changed execution conditions, not a measured F1 change or statistical stability bound. 

\subsection{Bounded Corpus Overlap}
Table~\ref{tab:overlap} reports exact processed-input overlaps within the selected records. Thirty-four test questions match \code{comp\_original} Q\&A records, and four supplied contexts match \code{banglahall\_context} records, yielding 38 distinct inputs (1.51\% of the test inputs). All 38 corresponding records are retained for their matching input in the final retrieval output. Among the 34 question-matched stored answers, only one is text-identical to the candidate response; answer disagreement does not establish that either answer is correct.

\begin{table}[t]
\caption{Bounded exact overlap against 2,516 saved processed inputs. Counts are matched inputs, not prediction errors or hidden-label matches.}
\label{tab:overlap}\centering\small
\begin{tabular}{p{0.56\columnwidth}rr}
\toprule
Selected collection/source & Records & Matches\\
\midrule
Q\&A / \code{comp\_original} & 77 & 34\\
Q\&A / \code{banglahall\_qna} & 1 & 0\\
Context / \code{comp\_original} & 88 & 0\\
Context / \code{banglahall\_context} & 924 & 4\\
\midrule
Total records / distinct matched inputs & 1,090 & 38\\
\bottomrule
\end{tabular}
\end{table}

Across all five retained slots per input (12,580 occurrences), these four source categories account for 816 occurrences (6.49\%), appearing on 729 inputs (28.97\%). This broader coverage includes nonexact retrieval matches. Retention is not proof of final judge consumption: a textual prompt may use only three records, and math paths use different selections. The source labels and exact matches establish overlap exposure, but not hidden-label leakage or a rules violation. Their provenance and permitted split membership remain unresolved. The bounded audit supplies no upper bound on overlap elsewhere in the corpus.

\section{Error Analysis and Operational Diagnostics}
\subsection{Observed Operational Failures}
The Version 2 MathAgent traces contain 204 program-run events, 49 of which carry stderr markers across 35 distinct examples. There are 49 forced-code-fix events and two textual fallback events, each after two execution attempts. Candidate checks are attempted on 96 rows, with 88 accepted terminal matches. These counts describe execution behavior; some errors are repaired, and successful execution may still yield an incorrect answer. A printed count of zero \code{step4\_error} stages would miss these events because that stage label is not an execution-failure counter.

One unresolved example, test ID 433, asks for a real-number inequality solution. Its saved log includes SymPy's \code{TypeError: cannot determine truth value of Relational}. The internal verdict remains unknown and becomes faithful under Eq.~\eqref{eq:default}. This is a verified operational failure and default assignment, not a verified classification error: the private reference label is unavailable. The same row is unresolved in both runs.

Several trace categories also describe policies rather than validated outcomes. The \code{step4\_math\_code} category contains both program-assisted decisions and fallback judgments, while \code{ctx+} records context presence rather than demonstrated use. Consequently, an error taxonomy based only on stage names can undercount failures or misidentify the governing evidence. The released analysis distinguishes event counts, unique affected examples, and terminal labels.

\subsection{Mechanistic Risks, Not Measured Ablations}
Response-conditioned retrieval can improve claim specificity in principle, but can also favor text supporting an erroneous candidate. Bare-question routes may counter this bias, yet no controlled ablation measures that effect. Likewise, a large judge may detect subtle inconsistencies, but these artifacts do not establish that 27B is a minimum viable size, that smaller judges collapse, or that forcing a verdict improves calibration. A format retry changes output coverage, not necessarily correctness.

The normalization allowlist and omitted MathAgent context create additional risks. Removal of foreign-script text, inequality symbols, or tag-like spans may alter the candidate claim; context-dependent reasoning can then lose a necessary premise at routing. Two saved processed responses are empty (IDs 633 and 665) and receive different verdicts. Without original-input comparison and an empty-response annotation policy, these are not established raw-data defects or classification errors. The batch and 27B workers left-truncate formatted prompts, potentially removing initial instructions or evidence; MathAgent truncation direction remains tokenizer-dependent. Such events are unlogged. Preserving original text and retained prompt spans is necessary for a future robustness evaluation.

The dominant textual-judge time suggests a practical target for future efficiency work, but does not justify replacing that model without accuracy measurements. Similarly, frequent candidate matching indicates reliance on the corpus, not independent generalization. A discriminating experiment would remove exact overlaps, compare question-only with response-conditioned retrieval, and disable the candidate shortcut under a fixed environment and independent labeled split. Those experiments were not performed here.

\section{Limitations, Reproducibility, and Responsible Use}
The strongest empirical limitation is the lack of gold labels for the saved evaluation inputs. We cannot assess per-class error trade-offs, paired significance, calibration, subgroup behavior, or component contributions. Published starter scores are not used as controlled baselines: they involve different validation subsets or leaderboard artifacts. A system report with a strong displayed score is not a substitute for an independent comparative study.

Corpus independence is a second limitation. The publicly uploaded IUFree package is not established as organizer-provided. It declares CC0 at the dataset level, but upstream attribution, licensing, collection dates, and split provenance are incomplete. The package contains benchmark-related source names and demonstrable exact overlaps. No conclusion about performance on unseen facts, languages, dialects, or later competition folds follows from the available result.

Reproducibility is partial rather than absent: exact executed notebooks, predictions, worker code, and timing metadata are available, enabling the analyses reported here. However, unpinned package/model revisions, incomplete index-building provenance, first-manifest discovery, and changed outputs across runs prevent a claim of bitwise inference reproducibility. Our supplementary scripts reproduce artifact statistics, not the private score or a newly trained model. A robust future release should pin checkpoint and corpus revisions, preserve input hashes, record truncation and source selection, and distinguish code execution from model-only fallback in telemetry.

Finally, unrestricted generated Python is unsuitable for deployment against untrusted inputs. A safer design requires a restricted mathematical interpreter or isolated execution service, resource limits, no network or secrets, and explicit source trust policies. Retrieved content can also be incorrect or adversarial. The faithful-class default must be disclosed to downstream users; it should not silently certify an unresolved answer in high-stakes use. These safeguards are recommendations, not features retroactively attributed to the competition implementation.

\section{Conclusion}
We described a retrieval-augmented and program-assisted Bengali hallucination detector and characterized its behavior through version-specific public artifacts. The score-linked Version 2 is associated with a displayed private macro-F1 of 0.956 and a 220.99-minute two-T4 execution. Saved outputs establish the distribution of textual and mathematical judgment paths, while traces reveal execution problems obscured by terminal stage labels. Cross-version differences and a bounded corpus-overlap audit qualify the interpretation of this result. The architecture offers a concrete way to combine retrieved evidence and computation, but stronger claims require corpus-controlled evaluation, labeled error analysis, controlled component comparisons, and a pinned, secure inference environment.

\bibliographystyle{IEEEtran}
\bibliography{references}
\end{document}